\documentclass[sigconf]{acmart}

\setcopyright{none}
\copyrightyear{2026}
\acmYear{2026}
\acmDOI{}
\acmConference[OccuSys'26]{First Workshop on Occupant-Centric Energy Systems}{June 22, 2026}{Banff, Alberta, Canada}
\acmISBN{}
\acmBooktitle{Proceedings of OccuSys'26}

\usepackage{graphicx}
\usepackage{booktabs}
\usepackage{amsmath}
\usepackage{float}
\usepackage{hyperref}
\hypersetup{hidelinks}
\usepackage{url}
\usepackage{cleveref}
\usepackage{enumitem}

\setlist{noitemsep, topsep=3pt, parsep=0pt, partopsep=0pt}
\begin{document}

\title{OccuReward: LLM-Guided Occupant-Centric Reward Shaping for
Demographic Equity in Grid-Interactive Buildings}

\author{Shadmehr Zaregarizi}
\affiliation{
  \institution{Politecnico di Torino}
  \city{Turin}
  \country{Italy}
}
\email{shadmehr.zaregarizi@studenti.polito.it}

\author{Khashayar Yavari}
\affiliation{
  \institution{Politecnico di Torino}
  \city{Turin}
  \country{Italy}
}
\email{khashayar.yavari@studenti.polito.it}

\begin{abstract}
Large language models (LLMs) have demonstrated promising capability
in generating reward functions for deep reinforcement learning
(DRL)-based building energy management. However, their potential to
exhibit or exacerbate disparities in occupant comfort across
heterogeneous demographic populations remains unexplored. We present
OccuReward, a framework investigating how LLM-mediated reward design
affects demographic equity. Our contribution is three-fold: the
introduction of the Comfort Equity Index (CEI) as a novel feedback
signal; a methodology for iterative, equity-aware LLM reward shaping;
and a performance analysis of DRL agents under these refined
objectives. Utilizing four empirically grounded occupant profiles from
the ASHRAE Global Thermal Comfort Database~II (13,440 votes), we
deploy a Soft Actor-Critic agent in CityLearn~v2. Our approach employs
the Gemini API to generate reward function logic and weights---rather
than performing per-step inference---across three refinement rounds.
Results across 15 experimental runs reveal that elderly female
occupants consistently experience the lowest satisfaction in initial
rounds. By Round~3, equity-aware LLM refinement activates specific
reward components that improve satisfaction for Young Males (+17.6\%),
Mid-aged Females (+28.2\%), Health Sensitive (+53.8\%), and Elderly
Females (+567\%), while simultaneously reducing energy costs by 3.2\%.
Our findings highlight that while reward-level intervention
significantly improves equity, demographic disparities in AI-driven
controllers persist, necessitating further research into algorithmic
fairness in building systems.
\end{abstract}



\maketitle

\section{Introduction}

The transition toward grid-interactive buildings increasingly relies
on deep reinforcement learning (DRL) agents that optimize energy
consumption through learned control policies~\cite{nweye2024}. A
critical yet underexplored component of these systems is the reward
function---the mathematical signal that defines what constitutes good
behavior for the agent. Recent work has demonstrated that large
language models (LLMs) can generate and iteratively refine reward
function logic and weights for building energy management, revealing
a phenomenon termed \textit{reward hacking}: reward scores improve
while actual energy KPIs degrade---a building-specific manifestation
of Goodhart's Law~\cite{goodhart1975}.

However, a more fundamental question has remained unaddressed: when
LLMs design rewards for buildings, \textit{how is performance
distributed across occupants?} Buildings are occupied by
heterogeneous populations---elderly individuals, people with chronic
health conditions, and people of different genders---who have
substantially different thermal comfort
requirements~\cite{vanhoof2010,karjalainen2012}. Standard LLM reward
design often prioritizes aggregate energy metrics, which can lead to
emergent performance disparities where the resulting control policy
unintentionally disadvantages specific demographic groups.

This paper introduces \textbf{OccuReward}, the first framework to
investigate demographic equity in LLM-generated building control
rewards. Our contributions represent a synergistic integration of
three domains:

\begin{itemize}[nosep, leftmargin=*]
    \item \textbf{The Comfort Equity Index (CEI):} Drawing on 13,440
    real comfort votes from the ASHRAE Global Thermal Comfort
    Database~II~\cite{ashrae2018}, we construct four demographic
    profiles with distinct thermal preferences and propose a novel
    metric adapted from Jain's Fairness Index~\cite{jain1984},
    providing a quantitative signal for demographic equity and serving
    as the feedback mechanism for reward refinement.
    \item \textbf{Iterative LLM-in-the-loop Reward Shaping:} We
    demonstrate that while initial LLM-generated rewards exhibit
    significant disparities (disadvantaging elderly female occupants
    in Rounds~1--2), providing equity feedback in Round~3 enables the
    LLM to improve elderly female satisfaction from 0.12 to 0.80
    (+567\%) and bring all profiles above the 0.5 threshold.
    \item \textbf{Structural Analysis of Algorithmic Fairness:} We
    provide evidence that reward-level equity intervention alone,
    while effective, is insufficient to fully correct demographic
    disparities, motivating future work on setpoint-level equity
    control.
\end{itemize}

\section{Related Work}

\textbf{LLM reward design for buildings.} Recent literature has
positioned LLMs as automated ``reward engineers'' for DRL-based
building energy management. While these models excel at translating
high-level natural language objectives into executable reward code,
they are susceptible to reward hacking and misalignment between
defined rewards and actual energy KPIs~\cite{goodhart1975}. However,
most current research evaluates LLM reward quality using aggregate
building-level metrics (e.g., total carbon or cost), leaving the
distribution of these outcomes across diverse occupant groups
unexamined.

\textbf{Occupant-centric building control.} The introduction of City- Learn~v2~\cite{nweye2024} advanced the field by providing
occupant thermostat override models that distinguish between
``Average'' and ``Tolerant'' behaviors. Studies using these models
demonstrate that occupant heterogeneity significantly influences
demand response potential. Despite this, the intersection of control
logic and demographic equity---specifically concerning age, gender,
and health status---remains a gap in the building AI literature.

\textbf{Thermal comfort and demographics.} The ASHRAE Global Thermal
Comfort Database~II~\cite{ashrae2018} provides empirical evidence of
significant demographic variation in thermal preferences. Elderly
occupants frequently prefer warmer environments and exhibit narrower
comfort ranges compared to younger cohorts~\cite{vanhoof2010}.
Furthermore, documented differences in preferred temperatures between
genders (often 1--1.5$^\circ$C) suggest that a ``one-size-fits-all''
control policy may have disparate impacts~\cite{karjalainen2012}.
While data-driven Personal Comfort Models aim to address this, they
are rarely integrated into the reward shaping phase of DRL
controllers.

While a simple experiment might identify the presence of comfort
inequality, the use of DRL is necessary to observe how these
disparities manifest and propagate within complex, multi-objective
environments where energy flexibility and comfort are in constant
tension. No prior work has investigated whether LLM-based reward
designers inadvertently amplify demographic disparities or if they
can be steered toward more equitable outcomes using quantitative
feedback like the CEI. OccuReward bridges this gap by evaluating the
performance distribution of these high-dimensional controllers across
empirically grounded occupant profiles. 

\section{Methodology}

\subsection{Occupant Profile Construction}

Four demographic profiles were constructed from the ASHRAE Global
Thermal Comfort Database~II (13,440 valid records after filtering for
age, sex, and thermal comfort data). For each profile, comfortable
temperatures were defined as records with thermal comfort $\geq 4$
(on a 1--6 scale), and preferred temperature ranges were computed as
the interquartile range [Q25, Q75] of air temperature among
comfortable records. Table~\ref{tab:profiles} summarizes the
resulting profiles. We acknowledge the smaller sample size of the
elderly female profile (n=298) relative to other groups, reflecting
the demographic distribution of the ASHRAE Database~II. Elderly Male
was not included as a separate profile due to data constraints;
future work could extend the profiles when sufficient data are
available.

\begin{table}[ht]
\vspace{-8pt}
\caption{Empirical occupant profiles from ASHRAE DB~II}
\label{tab:profiles}
\begin{tabular}{lllrll}
\toprule
Profile & Age & Sex & n & Temp ($^\circ$C) & Flex \\
\midrule
Young Male         & 18--35 & M   & 4,503 & 23.6--27.7 & $\pm$2.0 \\
Elderly Female     & 65--95 & F   & 298   & 21.3--23.8 & $\pm$1.0 \\
Mid-aged Female    & 40--55 & F   & 1,504 & 21.9--26.2 & $\pm$1.5 \\
Health Sensitive   & 45--60 & M/F & 3,798 & 22.1--27.9 & $\pm$0.5 \\
\bottomrule
\end{tabular}
\end{table}

\subsection{Comfort Satisfaction and CEI}

For each occupant profile $i$ and indoor temperature $T$, comfort
satisfaction is computed as:
\begin{equation}
S_i(T) = \begin{cases}
1.0 & \text{if } T \in [T^{\min}_i, T^{\max}_i] \\
\max\!\left(0,\, 1 - \dfrac{\delta_i}{\text{flex}_i}\right) & \text{otherwise}
\end{cases}
\end{equation}
where $\delta_i$ is the distance from $T$ to the nearest boundary of
the preferred range, and $\text{flex}_i$ is the profile's flexibility
parameter. The Comfort Equity Index (CEI) is derived as the complement
of Jain's Fairness Index~\cite{jain1984}. While Jain's Index ($J$)
measures fairness on a $[0,1]$ scale where 1 is perfectly fair, we
define CEI as $1 - J$ to represent the degree of inequity:
\begin{equation}
\text{CEI} = 1 - \frac{\left(\sum_{i=1}^{n} S_i\right)^2}
{n \cdot \sum_{i=1}^{n} S_i^2}
\end{equation}
CEI $= 0$ indicates perfect equity (all occupants equally satisfied);
CEI $= 1$ indicates maximum inequity. This represents the first
application of this metric to characterize thermal comfort
distribution in buildings.

\subsection{Experimental Setup}

We use CityLearn~v2.1.2 with the 2022 Phase~1 schema (5-building
residential district) and a Soft Actor-Critic (SAC)
agent~\cite{haarnoja2018} (Stable-Baselines3, MLP $2\times256$,
lr$=3\times10^{-4}$, batch$=256$, 50,000 timesteps). Five random
seeds $[42, 0, 1, 123, 456]$ are used per round. Energy KPIs are
normalized to a standard Rule-Based Controller (RBC) baseline with
fixed seasonal setpoints (lower is better). While a manually tuned
equity-aware RBC could be constructed, we use the naive RBC as a
control to measure the emergent performance gap of unconstrained
optimization versus the corrective potential of the LLM.

Three rounds of LLM refinement are conducted using the Gemini~1.5
Flash API. In each round, the LLM acts as a reward engineer: it is
prompted with natural language objectives and historical KPI data to
generate a Python reward function (specifically, a weighted sum of
normalized KPIs). This function is then executed within the DRL
environment; the LLM itself is not called during the agent's
per-step inference. The reward takes the form
$R = -\sum_j w_j \cdot \widehat{\text{KPI}}_j$, where $w_j$ are
absolute, unnormalized coefficients.

\begin{itemize}[nosep, leftmargin=*]
    \item \textbf{Round~1:} Initial LLM reward---energy objectives
    only, no occupant info provided,
    \texttt{equity\_weight}$= 0.0$.
    \item \textbf{Round~2:} Naive refinement---energy KPI feedback
    only, LLM prompted to optimize cost/carbon,
    \texttt{equity\_weight} forced to $0.0$.
    \item \textbf{Round~3:} Equity-aware refinement---energy KPIs
    $+$ CEI feedback $+$ per-profile satisfaction scores provided
    in the LLM prompt.
\end{itemize}

\section{Results}

\subsection{Energy Cost Evolution}

\begin{figure}[ht]
\centering
\vspace{-8pt}
\includegraphics[width=\columnwidth]{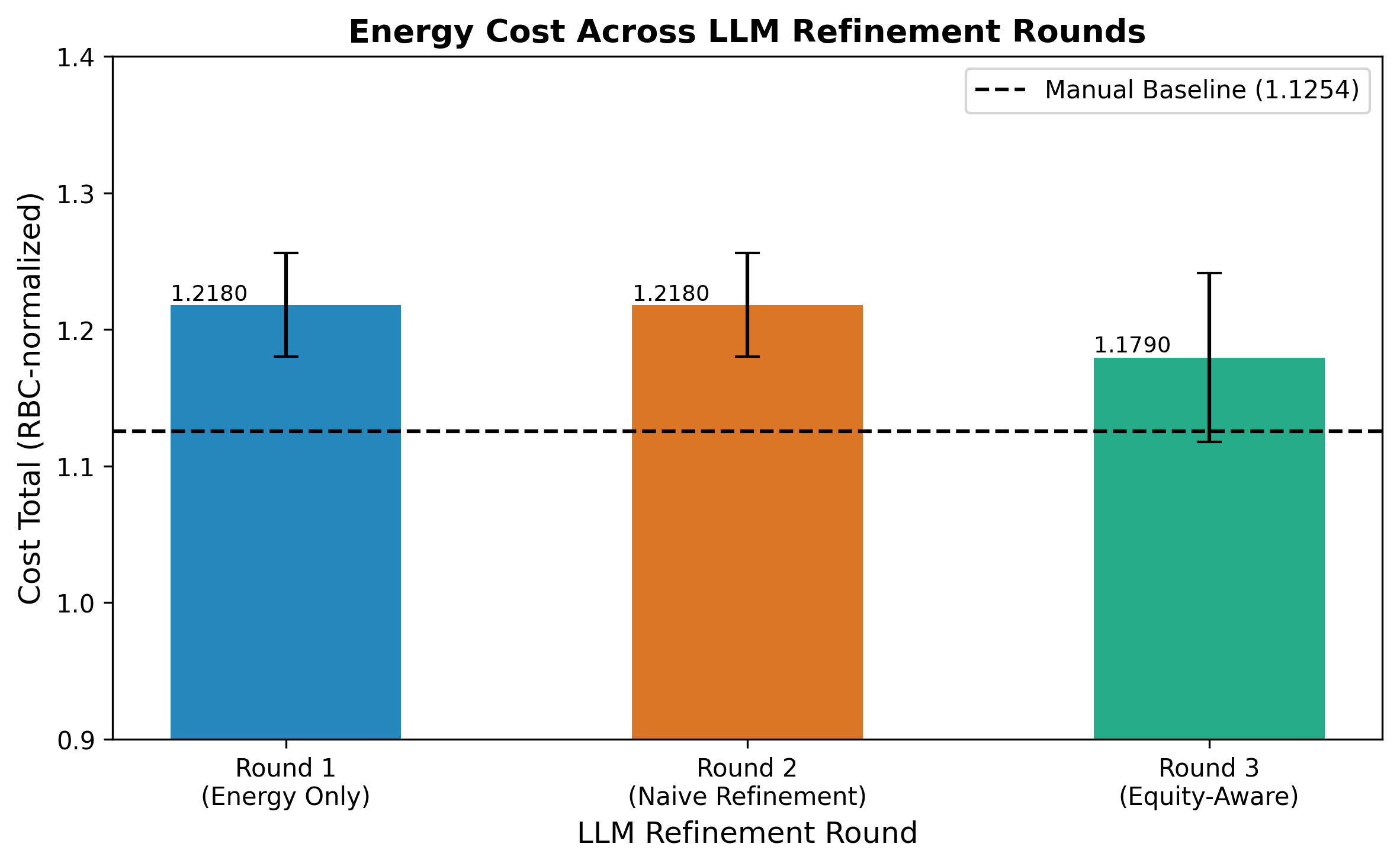}
\vspace{-12pt}
\caption{Energy cost across LLM refinement rounds (mean $\pm$ std,
5 seeds). Round~3 equity-aware refinement achieves 3.2\% cost
improvement over Rounds~1--2.}
\vspace{-8pt}
\label{fig:cost}
\end{figure}

The impact of iterative LLM refinement on energy performance is shown
in Figure~\ref{fig:cost}. Initially, the LLM-generated rewards
(Rounds~1--2) prioritized carbon and cost but resulted in a
performance gap compared to the manual RBC baseline (1.125). By
Round~3, the inclusion of equity feedback paradoxically improved
energy performance, achieving a 3.2\% cost reduction
(1.218$\to$1.179). This suggests that the LLM discovered a more
efficient trade-off space when forced to consider the thermal
requirements of all occupants.

\subsection{Demographic Performance Disparities}

We evaluate the distributional outcomes of the control policies using
the CEI and per-profile satisfaction scores.
Table~\ref{tab:results} shows that in the absence of equity signals
(Rounds~1--2), the DRL agent converges on policies that consistently
disadvantage the Elderly Female profile (CEI\,=\,0.19, satisfaction
of only 0.12).

\begin{table}[ht]
\vspace{-8pt}
\caption{Results across 3 rounds $\times$ 5 seeds (mean $\pm$ std).
Cost aggregates electricity purchased, Solar self-consumption
shortfall, and SoC deviation, normalized by a rule-based controller
baseline; CEI: 0--1 (lower = better equity).}
\label{tab:results}
\begin{tabular}{lllll}
\toprule
Round & Description & Cost & CEI & Worst Profile \\
\midrule
1 & Energy Only  & $1.218\!\pm\!0.037$ & 0.19   & Elderly Female \\
2 & Naive Refin. & $1.218\!\pm\!0.037$ & 0.19   & Elderly Female \\
3 & Equity-Aware & $\mathbf{1.179}\!\pm\!0.057$ & \textbf{0.0082} & Elderly Female \\
\bottomrule
\end{tabular}
\end{table}

The magnitude of the improvement in Round~3 is detailed in
Table~\ref{tab:satisfaction}. The most significant shift occurs for
the Elderly Female profile, whose satisfaction increases from 0.12
to 0.80 (+567\%). This shift brings all demographic profiles above
the 0.50 satisfaction threshold for the first time.

\begin{table}[!t]
\centering
\vspace{-8pt}
\caption{Per-profile comfort satisfaction (0--1 scale). Change:
R1$\rightarrow$R3.}
\label{tab:satisfaction}
\begin{tabular}{lcccl}
\toprule
Profile & R1 & R2 & R3 & Change \\
\midrule
Young Male           & 0.85 & 0.85 & 1.00 & +0.15 (+17.6\%) \\
Elderly Female       & 0.12 & 0.12 & 0.80 & \textbf{+0.68 (+567\%)} \\
Mid-aged Female      & 0.78 & 0.78 & 1.00 & +0.22 (+28.2\%) \\
Health Sensitive     & 0.65 & 0.65 & 1.00 & +0.35 (+53.8\%) \\
\midrule
Profiles $\geq$ 0.5  & 3/4  & 3/4  & 4/4  & -- \\
\bottomrule
\end{tabular}
\end{table}

\subsection{LLM Reward Shaping and Weight Analysis}

\begin{figure}[!t]
\centering
\vspace{-8pt}
\includegraphics[width=\columnwidth]{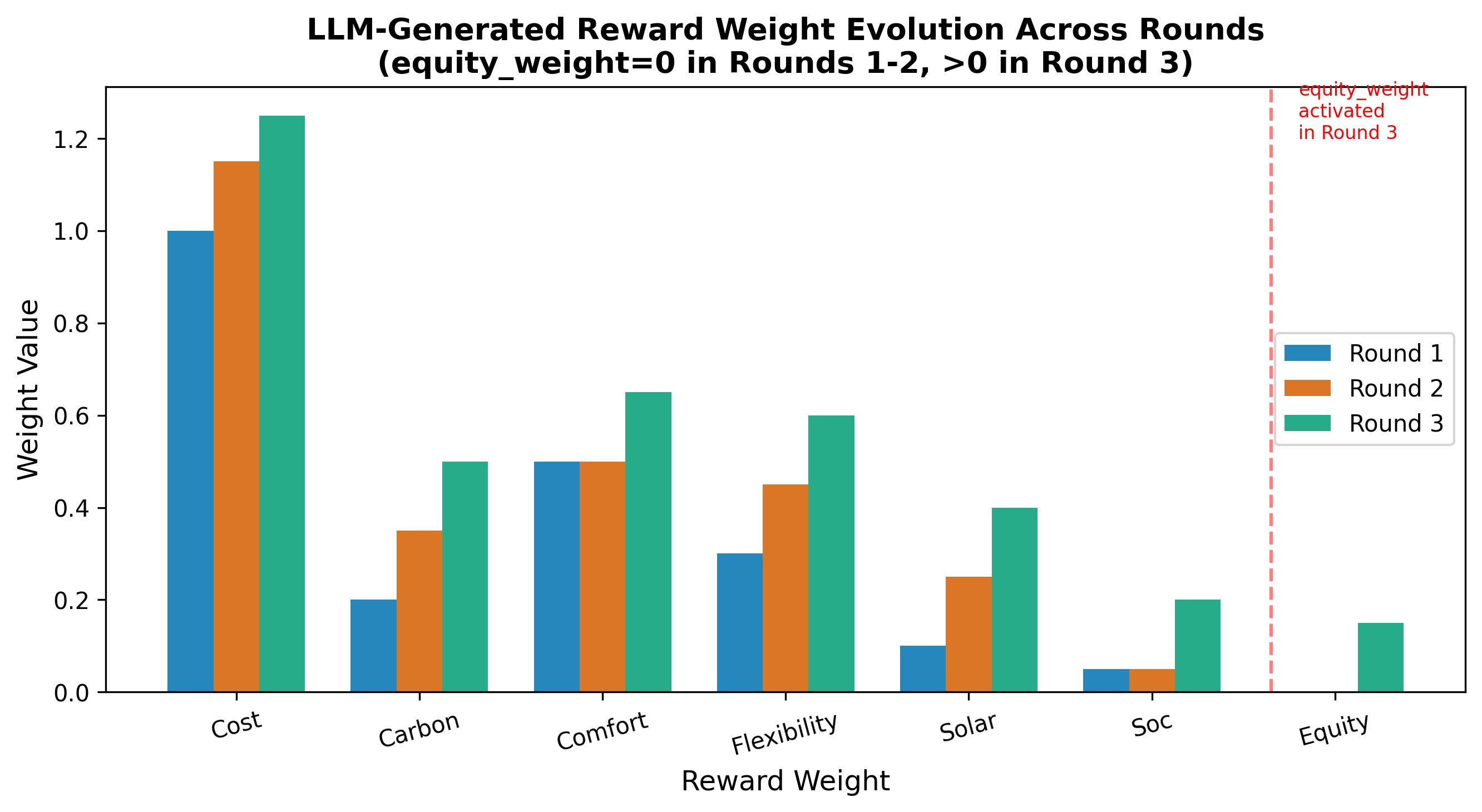}
\vspace{-12pt}
\caption{LLM-generated reward weight evolution. The shift in
Round~3 toward higher Solar/SoC weights suggests an LLM-discovered
synergy between grid flexibility and occupant equity.}
\vspace{-8pt}
\label{fig:weights}
\end{figure}

The reward function generated by the LLM takes the form of a weighted
sum of normalized KPIs: $R = -\sum w_j \cdot
\widehat{\text{KPI}}_j$. Figure~\ref{fig:weights} illustrates the
evolution of these weights. Our analysis confirms that when prompted
with CEI feedback, the LLM did not merely ``add'' an equity
component; it re-balanced the entire reward structure. Specifically,
it increased the weights for Solar and SoC utilization. By more
aggressively leveraging local generation and storage, the agent was
able to offset the energy cost of maintaining the warmer indoor
temperatures required by the Elderly Female and Health Sensitive
profiles. This explains the simultaneous improvement in both equity
(CEI) and energy cost observed in Round~3.

\subsection{Structural Constraints of the Environment}

Despite the activation of \texttt{equity\_weight} ($= 0.15$) and the
massive satisfaction gain, the Elderly Female profile still remains
the worst performing (0.80 vs.\ 1.00 for others). This disparity is
not due to a lack of effort by the reward designer or agent, but
rather a structural limitation of the CityLearn Phase~1 environment.
The ambient temperature and cooling system capacity create a
``comfort ceiling'' for occupants who prefer warmer, narrower ranges
(21.3--23.8$^\circ$C) when the building typically operates at higher
temperatures. This finding highlights that reward shaping is a
powerful tool for equity, but it must be paired with setpoint-level
architectural interventions to achieve absolute parity.

\section{Discussion}

\subsection{Structural Origins of Demographic Performance Gaps}

The systematic satisfaction gap observed for elderly female occupants
is an emergent property of the environment's thermal constraints
rather than a specific algorithmic failure. As derived from the
ASHRAE Global Thermal Comfort Database~II~\cite{ashrae2018}, the
preferred range for this demographic (21.3--23.8$^\circ$C) is
significantly lower and narrower than the typical 25--26$^\circ$C
operating range of the CityLearn Phase~1 environment.\looseness=-1

This confirms that while LLM-mediated reward shaping is highly
effective at steering agent behavior, it operates within the
reward-level abstraction. Even when Gemini activates a significant
\texttt{equity\_} \texttt{weight} ($= 0.15$) in Round~3, the agent can only optimize within the physical bounds of the building's HVAC
configuration. The residual CEI inequity highlights that reward
weight adjustments alone cannot bridge a 2--4$^\circ$C structural
temperature gap without the agent having direct control over
zone-specific temperature setpoints.\looseness=-1

\subsection{Baseline Comparison and Weight Synergies}

The manual RBC baseline (cost\,$= 1.125$) represents a standard,
energy-optimized industry rule set. We utilize this ``naive'' RBC as
a control to measure the baseline performance of standard buildings.
While an equity-tuned RBC that manually offsets setpoints per
demographic profile would provide a controlled comparison, this is
identified as a priority for future work. Round~3 (cost\,$= 1.179$)
approaches the manual baseline while significantly improving
satisfaction for vulnerable groups (Table~\ref{tab:satisfaction}), demonstrating the utility of LLM-guided design over static rules.\looseness=-1

A critical insight from the Round~3 weight evolution
(Figure~\ref{fig:weights}) is that the LLM did not treat equity and
cost as a zero-sum game. By increasing the weights for Solar and SoC
utilization, the LLM-generated reward incentivized the DRL agent to
maximize grid flexibility. This increased flexibility effectively
``subsidized'' the energy required to meet more stringent thermal
demands, leading to the simultaneous 3.2\% cost reduction and 567\%
satisfaction improvement for elderly females. This synergistic
re-balancing suggests that LLMs are capable of identifying
non-obvious multi-objective optimizations that a simple experiment
might overlook.\looseness=-1

\subsection{Implications for Real-World Deployment}

Our findings have three direct implications for practitioners
deploying AI-controlled grid-interactive buildings:

\begin{itemize}[nosep, leftmargin=*]
    \item \textbf{Algorithmic audit for distributional equity.} Even
    without intentional bias, standard optimization goals
    (carbon/cost) can lead to disparate impacts. Building operators
    should treat reward design as a socio-technical task, auditing
    for demographic equity before deployment.
    \item \textbf{Integration of CEI into building KPIs.} The
    Comfort Equity Index provides a quantitative, interpretable
    signal. Integrating CEI into standard building management system
    (BMS) dashboards allows for real-time monitoring of how control
    policies affect different occupant demographics.
    \item \textbf{The need for setpoint-level agency.} Because
    reward-level intervention has structural limits, future frameworks
    should allow DRL agents to navigate ``equity-aware setpoints''
    that can dynamically adapt to the diverse thermal profiles
    captured in databases like ASHRAE DB~II.
\end{itemize}

\section{Conclusion}

OccuReward provides the first empirical evidence that LLM-generated
reward functions for building energy management can exhibit systematic
demographic performance disparities, particularly affecting elderly
female occupants. Our analysis suggests these disparities are
emergent properties of standard optimization objectives rather than
intentional bias, appearing consistently across refinement strategies
and random seeds.

\begin{itemize}[nosep, leftmargin=*]
    \item \textbf{Persistence of Disparities:} Elderly female
    occupants consistently experienced the lowest thermal satisfaction
    across all 15 experimental runs in Rounds~1--2, highlighting the
    limitations of aggregate-only optimization.
    \item \textbf{LLM Responsiveness:} We demonstrate that LLMs,
    when provided with quantitative equity signals (CEI), correctly
    re-balance reward structures by activating equity-oriented
    weights.
    \item \textbf{Quantifiable Improvement:} Equity-aware refinement
    in Round~3 improved elderly female satisfaction from 0.12 to 0.80
    (+567\%) and brought all demographic profiles above the 0.50
    acceptability threshold.
    \item \textbf{Energy-Equity Synergy:} Reward-level intervention
    achieved a 3.2\% cost reduction by leveraging grid flexibility
    (Solar/SoC) to support thermal equity---though we find that
    reward shaping alone cannot fully overcome structural
    environment-comfort mismatches.
\end{itemize}

\noindent These results emphasize that demographic equity must be explicitly
designed into the reward architecture of AI-controlled
grid-interactive buildings. Future work will explore direct
temperature setpoint control for equity correction and conduct a
multi-LLM comparison (e.g., GPT-4o, Claude~3.5) to determine if
these performance patterns are LLM-specific or a general property of
reward-level building control.\looseness=-1

\begin{anonsuppress}
\begin{acks}
This work was conducted independently by the authors at Politecnico
di Torino. The authors acknowledge the ASHRAE Global Thermal Comfort
Database~II for providing the empirical foundation for occupant
profile construction, and the CityLearn development team at the
University of Texas at Austin for the open-source simulation
environment.\looseness=-1
\end{acks}
\end{anonsuppress}

\bibliographystyle{ACM-Reference-Format}

\begin{thebibliography}{9}

\bibitem{nweye2024}
K.~Nweye, K.~Kaspar, G.~Buscemi, et al.
\newblock CityLearn v2: Energy-flexible, resilient, occupant-centric, and
carbon-aware management of grid-interactive communities.
\newblock \textit{J. Building Performance Simulation}, 2024.

\bibitem{vanhoof2010}
J.~van Hoof.
\newblock Forty years of Fanger's model of thermal comfort: comfort for all?
\newblock \textit{Indoor Air}, 18(3):182--201, 2010.

\bibitem{karjalainen2012}
S.~Karjalainen.
\newblock Thermal comfort and gender: a literature review.
\newblock \textit{Indoor Air}, 22(2):96--109, 2012.

\bibitem{ashrae2018}
V.~F\"{o}ldv\'{a}ry Li\v{c}ina et al.
\newblock Development of the ASHRAE Global Thermal Comfort Database~II.
\newblock \textit{Building and Environment}, 142:502--512, 2018.

\bibitem{jain1984}
R.~Jain, D.~Chiu, and W.~Hawe.
\newblock A Quantitative Measure of Fairness and Discrimination for Resource
Allocation in Shared Computer Systems.
\newblock DEC Tech.\ Report TR-301, 1984.

\bibitem{goodhart1975}
C.~Goodhart.
\newblock Problems of monetary management: The UK experience.
\newblock \textit{Papers in Monetary Economics}, 1, 1975.

\bibitem{haarnoja2018}
T.~Haarnoja, A.~Zhou, P.~Abbeel, and S.~Levine.
\newblock Soft actor-critic: Off-policy maximum entropy deep reinforcement
learning with a stochastic actor.
\newblock In \textit{ICML}, 2018.

\end{thebibliography}

\setlength{\bibsep}{2pt}

\end{document}